\documentclass[times,twocolumn,final,authoryear]{elsarticle}

\usepackage{prletters}
\usepackage{framed,multirow}

\usepackage{amssymb}
\usepackage{latexsym}

\usepackage{url}
\usepackage{xcolor}
\usepackage{subfig}
\usepackage{multirow}

\definecolor{newcolor}{rgb}{.8,.349,.1}

\journal{Pattern Recognition Letters}

\begin{document}

\thispagestyle{empty}

\clearpage
\thispagestyle{empty}
\ifpreprint
  \vspace*{-1pc}
\fi






\clearpage
\thispagestyle{empty}

\ifpreprint
  \vspace*{-1pc}
\else
\fi

\begin{table*}[!t]
\ifpreprint\else\vspace*{-15pc}\fi

\section*{Research Highlights (Required)}

To create your highlights, please type the highlights against each
\verb+\item+ command. 

\vskip1pc

\fboxsep=6pt
\fbox{
\begin{minipage}{.95\textwidth}
It should be short collection of bullet points that convey the core
findings of the article. It should include 3 to 5 bullet points
(maximum 85 characters, including spaces, per bullet point.)  
\vskip1pc
\begin{itemize}

 \item We designed a novel architecture for video object detection that capitalizes on temporal information.

 \item We designed a novel fusion module to merge feature maps coming from several temporally close frames. 
 
 \item We proposed an improvement to the SpotNet attention module. 

 \item We trained and evaluated our architecture with three different base detectors on two traffic surveillance datasets.

 \item We demonstrated a consistent and significant improvement of our model over the three baselines.

\end{itemize}
\vskip1pc
\end{minipage}
}

\end{table*}

\clearpage

\ifpreprint
  \setcounter{page}{1}
\else
  \setcounter{page}{1}
\fi

\begin{frontmatter}

\title{FFAVOD: Feature Fusion Architecture for Video Object Detection}

\author[1]{Hughes \snm{Perreault}} 
\ead{hughes.perreault@polymtl.ca}
\author[1]{Guillaume-Alexandre \snm{Bilodeau}}
\author[1]{Nicolas \snm{Saunier}}
\author[2]{Maguelonne \snm{Héritier}}

\address[1]{Polytechnique Montréal, 2500 Chemin de Polytechnique, Montréal H3T 1J4, Canada}
\address[2]{Genetec, 2280 Boulevard Alfred Nobel, Montréal H4S 2A4, Canada}

\received{1 May 2013}
\finalform{10 May 2013}
\accepted{13 May 2013}
\availableonline{15 May 2013}
\communicated{S. Sarkar}

\begin{abstract}
A significant amount of redundancy exists between consecutive frames of a video. Object detectors typically produce detections for one image at a time, without any capabilities for taking advantage of this redundancy. Meanwhile, many applications for object detection work with videos, including intelligent transportation systems, advanced driver assistance systems and video surveillance. Our work aims at taking advantage of the similarity between video frames to produce better detections. We propose FFAVOD, standing for feature fusion architecture for video object detection. We first introduce a novel video object detection architecture that allows a network to share feature maps between nearby frames. Second, we propose a feature fusion module that learns to merge feature maps to enhance them. We show that using the proposed architecture and the fusion module can improve the performance of three base object detectors on two object detection benchmarks containing sequences of moving road users. Additionally, to further increase performance, we propose an improvement to the SpotNet attention module. Using our architecture on the improved SpotNet detector, we obtain the state-of-the-art performance on the UA-DETRAC public benchmark as well as on the UAVDT dataset. Code is available at \url{https://github.com/hu64/FFAVOD}.
\end{abstract}

\begin{keyword}
\MSC 41A05\sep 41A10\sep 65D05\sep 65D17
\KWD Video object detection\sep Feature fusion\sep Traffic scenes

\end{keyword}

\end{frontmatter}


\section{Introduction}
Object detection as meant in this paper is the task of finding rectangular bounding boxes that tightly bound objects of interest in an image. Video object detection uses the temporal information of a video in order to gain an edge over single frame detection. Object detection in general has been largely dominated by deep learning approaches in the last few years. It is quite a vibrant subject that has seen a lot of research. Its cousin task, video object detection, has seen a lot less activity on the other hand. To capitalize on video information, some methods aim to speed up the inference process by estimating feature maps~\citep{Liu_2018_CVPR}, and some aim to merge feature maps by optical flow warping~\citep{zhu2017flow}. In contrast, in this paper, we develop an end-to-end architecture and train it to merge feature maps without external methods or explicit knowledge of temporal relations. 

The applications for video object detection are certainly not lacking, for instance in intelligent transportation systems, such as advanced driver assistance systems and video surveillance. In a world where robotics and automation are growing, the number of these applications can only increase. 

\begin{figure}[t!]%
    \centering
    \includegraphics[width=0.8\linewidth]{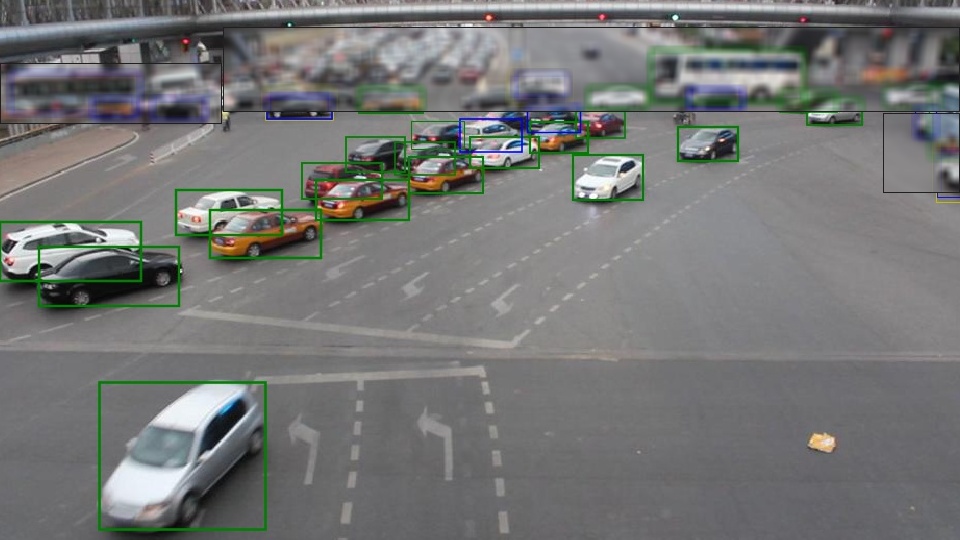}
    \caption{In this image from the UA-DETRAC dataset~\citep{Wen2015Tracking}, one can see multiple examples where the proposed architecture applied on SpotNet~\citep{perreault2020spotnet} (blue) outperforms its baseline (yellow). Detections by both models are shown in green (considering a minimum IOU score of 0.8 to match them). Examples of improved performance include cases of occlusion and of smaller objects (top center and both top corners). The blurred rectangles represent regions excluded for the evaluation, but where the proposed architecture nonetheless can make better detections.}
    \label{challenges}%
\vspace{-1em}
\end{figure}

Using multiple frames can improve results in situations that include occlusions, blur and smaller objects (see in figure~\ref{challenges}). Indeed, there are several ways in which using multiple frames can help to detect objects on a target frame.  In a video sequence, the objects of interest appear over and over again under different lighting, angles and occlusion conditions. Over a number of consecutive frames, one of them will contain the best view of an object of interest. One can even go even further, for each pixel location in our target frame, one of the temporally close frames will contain the best object features for this location. Of course, the features cannot be too far temporally because the object of interest might have moved too much in that case. FFAVOD aims to learn an operation to best merge the feature maps of several frames, at each image location. 

We propose three main contributions: an adaptable video object detection architecture that can be inserted into multiple object detection methods, a module for the fusion of feature maps in order to enrich them, and an improvement to the SpotNet~\citep{perreault2020spotnet} attention module. The result of these contributions is a framework called FFAVOD, which is trainable end-to-end for video object detection and classification. This work generalizes the work published in~\citep{perreault2020rn} by extending it and testing it with several baseline object detector architectures. 

The evaluation of our method is focused on traffic surveillance scenes, since they contain most of the challenges we aim to solve with our method and they are our target application. The method is evaluated on two datasets, UA-DETRAC~\citep{Wen2015Tracking} and UAVDT~\citep{du2018unmanned}. We tested our method by incorporating it into several base networks and comparing them to their corresponding baselines and other state-of-the-art methods on each dataset. A consistent and significant improvement over the base networks is demonstrated, as well as strong overall results on both datasets. 

\section{Related Work}
\subsection{Object Detection}
The object detection benchmarks have been systematically dominated by deep learning-based methods in the last few years. They can be broadly divided in three categories: two-stage, single-stage and anchor-free. 

The two-stage category contains methods that use an object proposal phase where object candidates are proposed, and then refined into final predictions. R-CNN~\citep{rcnn_Girshick_2014_CVPR} is the first dominant detector to make use of a CNN. However, the object proposal method it uses is an external one, selective search. The CNN is used to extract features for every object proposal. Those features are then classified using an SVM. Fast R-CNN~\citep{Girshick_2015_ICCV_fast} improves upon it by passing the image into a CNN only once, and then cropping the features from the resulting feature map for the corresponding region of the image for each object proposal. The third iteration of the method, Faster R-CNN~\citep{ren2015faster}, removes the external object proposal method. It does so by using two CNNs, one that proposes object candidates called the region proposal network (RPN), and another that classifies and refines the bounding box for each candidate. The two CNNs share most parameters, making Faster R-CNN very efficient with accurate results. R-FCN~\citep{RFCN_NIPS2016_6465} is a variant of Faster R-CNN, where the objects are detected as a grid of parts of objects where each cell of the grid votes, making it better for accurately positioning objects. The method Evolving Boxes~\citep{EB_wang2017evolving} is an efficient vehicle detection network that includes a proposal sub-network and an early discard sub-network. It generates candidates with multiple representations and later refines and classifies those candidates. 

The single-stage category improves over the two-stage methods by speeding up the inference process to eventually arrive to real-time detection. Single-stage refers to the removal of the object proposal phase. The original method YOLO~\citep{redmon2016you_yolo} is the first CNN-based method to reach real-time speed. It divides the image into a grid, and makes each cell predict two bounding boxes using regression. Two subsequent versions of YOLO were proposed, YOLOv2~\citep{redmon2016you_yolo} and YOLOv3~\citep{redmon2018yolov3}, with various improvements. SSD~\citep{liu2016ssd} addresses the challenge of detecting objects at multiple scales by using feature maps at multiple levels in the CNN. A sliding window approach is then used to perform classification and regression with anchor boxes at fixed aspect ratios and scales. RetinaNet~\citep{lin2018focal} is an improvement upon SSD that uses a different loss function, the focal loss, that aims to fix the asymmetry between positive and negative examples during training. RetinaNet also incorporates a feature pyramid network (FPN)~\citep{Lin_2017_CVPR_FPN}, a network that builds a pyramid of features at different scales by using lateral and vertical connections on feature maps of the CNN. Using a sliding window, RetinaNet then classifies and regresses on these pyramid levels using anchor boxes.   

More recently, a different approach to object detection was proposed. It is called anchor-free since it replaces the use for anchor boxes by detecting objects as keypoints. CornerNet~\citep{law2018cornernet} first introduced this approach by detecting objects as a pair of keypoints, the top-left and the bottom-right corners. A learned embedding allows the method to later pair the corresponding corners using the similarity between the embedding vectors. CenterNet (keypoint triplets)~\citep{duan2019centernet} builds upon this idea by adding a third learned keypoint, the center of the object, which is used to remove false positives, since two corners without a center are not likely to be part of an object. CenterNet (objects as points)~\citep{zhou2019objectsaspoints} uses a different approach, and instead trains a network to detect objects as a single center keypoint, using center heatmaps for each label, as well as regressions for the width and height and the offset. A variant of CenterNet (objects as points), SpotNet~\citep{perreault2020spotnet}, makes use of semi-supervised segmentation annotations to train a self-attention mechanism within the network and thus increase its performance. 

\subsection{Video Object Detection}
A first way to perform video object detection is to combine the features of several frames. Flow Guided Feature Aggregation (FGFA)~\citep{zhu2017flow} makes use of optical flow warping to merge temporally close frames to improve accuracy. In MANet~\citep{wang2018fully}, an optical flow estimation is done and two networks are trained, one to do pixel-level calibration (detailed motion adjustments) and another one for instance-level calibration (global motion adjustments). Some works take advantage of recurrent neural networks, for example STMM~\citep{xiao2018video} that models the motion and appearance of an object within a video sequence. In~\citep{Liu_2018_CVPR}, Long Short-Term Memories (LSTMs) are used to interpolate feature maps, which increases the inference speed greatly. Multi-frame Single Shot Detector~\citep{broad2018recurrent_mf-ssd} builds upon SSD by adding a temporal information with a recurrent convolutional module. ~\citep{Bertasius_2018_ECCV} used deformable convolutions to compute offsets between temporally close frames. Using these offsets they can share some features from neighboring frames to better perform detection. 3D-DETNet~\citep{3D_detnet_li20183d} makes use of 3D convolutions on temporally close frames that are concatenated in order to produce better feature maps. ~\citep{perreault2019road} experimented with training networks on image pairs. Since no pre-trained weights were available, they could only outperform the single frame baseline when training from scratch. The same problem is faced when using 3D convolutions.  
In contrast to these previous methods, we also merge feature maps, but instead of aligning the feature maps or using 3D convolutions, we train a network to fuse directly the raw features maps from several frames. 

 Another possible avenue is to combine detection and tracking. TrackNet~\citep{li2019tracknet} extends the Faster R-CNN framework by directly detecting a 3D cube bounding a moving object. In Joint detection and tracking in videos with identification features~\citep{munjal2020joint}, the authors train a multi-task model by joint optimization of detection, tracking and re-identification. The Global Correlation Network~\citep{lin2021global} also jointly trains the detection and the tracking task by first training the detection module before fine-tuning the whole network. Finally, motion information can be integrated to the network. Illuminating Vehicles With Motion Priors For Surveillance Vehicle Detection~\citep{wang2020illuminating} integrates motion in a network in order the illuminate the vehicles and suppress false positives. To better detect tiny objects, MMA~\citep{hu2019mma} proposes a dual stream network, an appearance stream and a motion stream. They also integrate a memory attention module to help select discriminating features using the temporal information. MFMNet~\citep{liu2019perceiving} uses a motion from memory module to encode the temporal context. This module contains a separate dynamic memory for every input sequence, and  produces motion features for every frame.  

\subsection{Feature Fusion Strategies}

Since we consider combining the features of several frames, we also briefly review common feature fusion strategies. Most methods consider the concatenation of features or their sum. In the inception network~\citep{szegedy2015going}, convolutions with kernels of various sizes are used to produce a set of feature maps which are then combined using concatenation. In the feature pyramid network~\citep{Lin_2017_CVPR_FPN}, pyramid levels are combined using upsampling and addition, using a top-down approach. In FSSD~\citep{li2017fssd}, a variant of the feature pyramid is proposed where feature maps from all levels are first concatenated together, and later used to create all the pyramid levels. In ExFuse~\citep{zhang2018exfuse}, low-level and high-level semantic information are merged after introducing spatial information into the high-level features and semantic information into the low-level features. The feature maps are combined using additions. 

\section{Proposed Method}
The task we aim to solve can be described as placing a bounding box and label on every object of interest in a target image, using the target image as well as $n$ frames both before and after the target frame. To do so, two main contributions are presented. First, we design an architecture for object detection that allows the use of temporally close frames and that is adaptable to multiple base networks. Second, we propose a fusion module to merge feature maps of the same dimension from temporally close frames. In order to achieve high accuracy with less training, we specifically designed this architecture to use pre-trained weights from state-of-the-art single frame methods. In order to achieve state-of-the-art detection results, we also propose an improvement to the SpotNet attention module.  

\subsection{Frame Fusion Architecture}
The idea behind our architecture is to compute feature maps for frames in a video only once for each frame and, when performing detection for a target frame, use the feature maps that are already computed for the temporally close frames to enhance the target frame feature map. Thanks to this idea of computing once and reusing, we also propose a novel fusion module that is used deep in the base network. We do not have to merge early in order the save computation time since we reuse the feature maps for detecting in the next frames. It is inserted between the backbone and the regression and classification heads of a base object detection network.

\begin{figure*}[t]
\begin{center}
\includegraphics[width=0.7\linewidth]{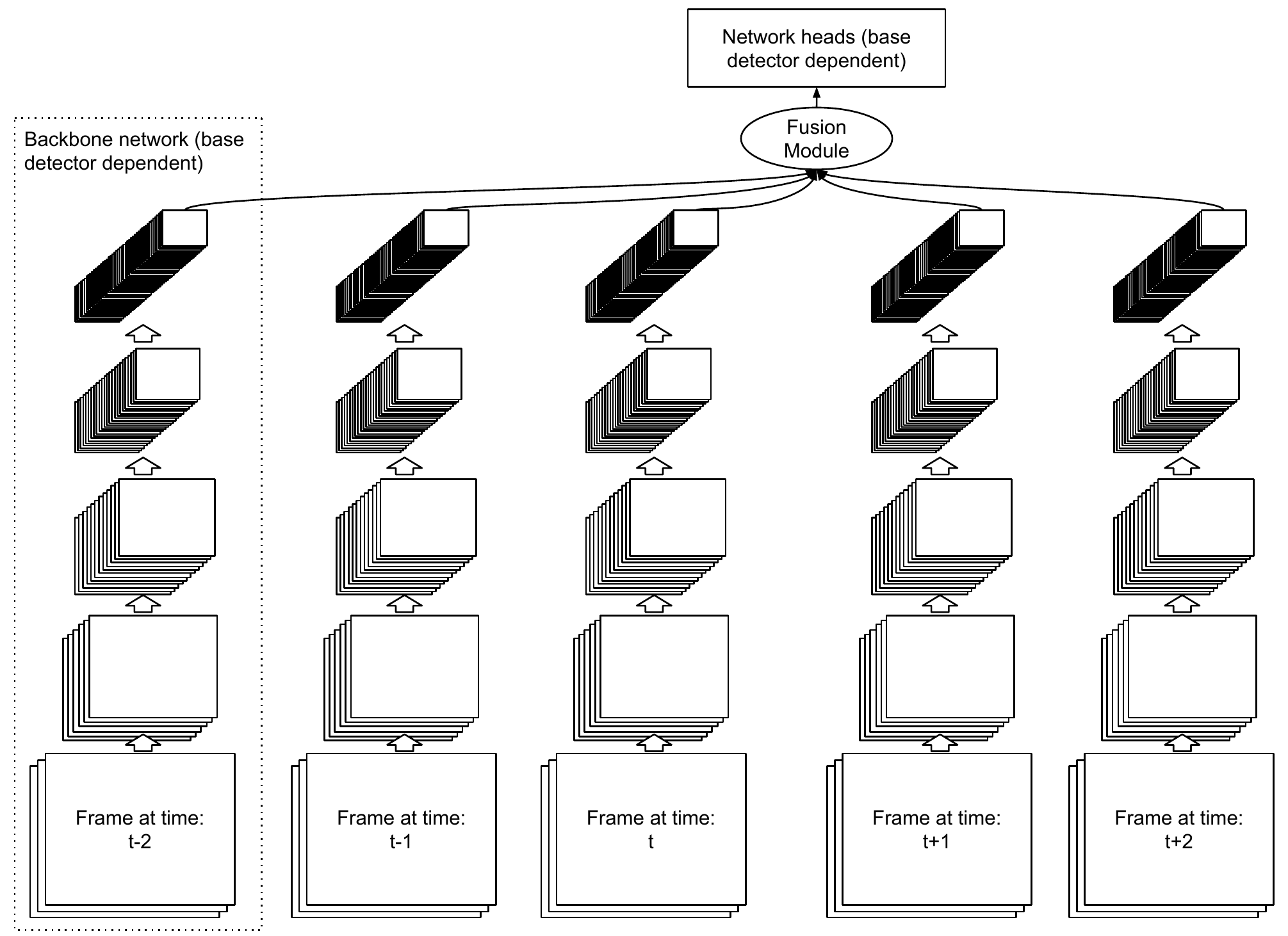}
\end{center}
\vspace{-1em}
   \caption{A visual representation of FFAVOD with a window of 5 frames ($n=2$). Frames are passed through the backbone network of the base object detection network, and the fusion module takes their outputs as input. Finally, the fusion module outputs a fused feature map compatible with the base object detection network, and the base object detection heads are applied to the fused feature map to classify the object categories and regress the bounding boxes.}
\label{architecture}
\end{figure*}

Our frame fusion architecture for video object detection (FFAVOD) takes multiple images as input, and they are merged as one feature map deeper in the network, as shown in figure~\ref{architecture}. For a target frame at time $t$, we use a window of $2n + 1$ frames, that is $n$ frames before and after the frame $t$. We cannot use a $n$ that is too high because the positions of objects of interest at the different frames will become too different and it will not be possible to share the information. When facing ``boundary conditions'', meaning that we are too close to the beginning or end of a sequence to take frame $t-n$ or $t+n$ for instance, we simply duplicate the first or last frame. For example, not having access to $t-2$ but to $t-1$, with $n = 2$, we would use the five frames $t-1$, $t-1$, $t$, $t+1$ and $t+2$. 

The way feature maps are merged is adapted to each base detector. For example, for a network like RetinaNet, the three outputs used to create the feature pyramid network are merged. For networks like CenterNet, the outputs of the double stacked hourglass network used as backbone are merged. The merged feature map is used to create the center keypoint heatmaps. Details about which layers are merged are provided for three networks in section~\ref{basedet}.

The fusion process is done with our custom fusion module described below, that enhances the target frame feature map. During the inference process, we can reuse feature maps already computed as we progress sequentially in the video, thus saving time for computing detections at every frame. However, during the training process, multiple images must e used for every ground-truth example, thus requiring more memory and time to train. We do believe that the extra time required to train the model is worth the better detection performance. 

\subsection{Fusion Module}
\begin{figure}[t]
\begin{center}
\includegraphics[width=0.8\linewidth]{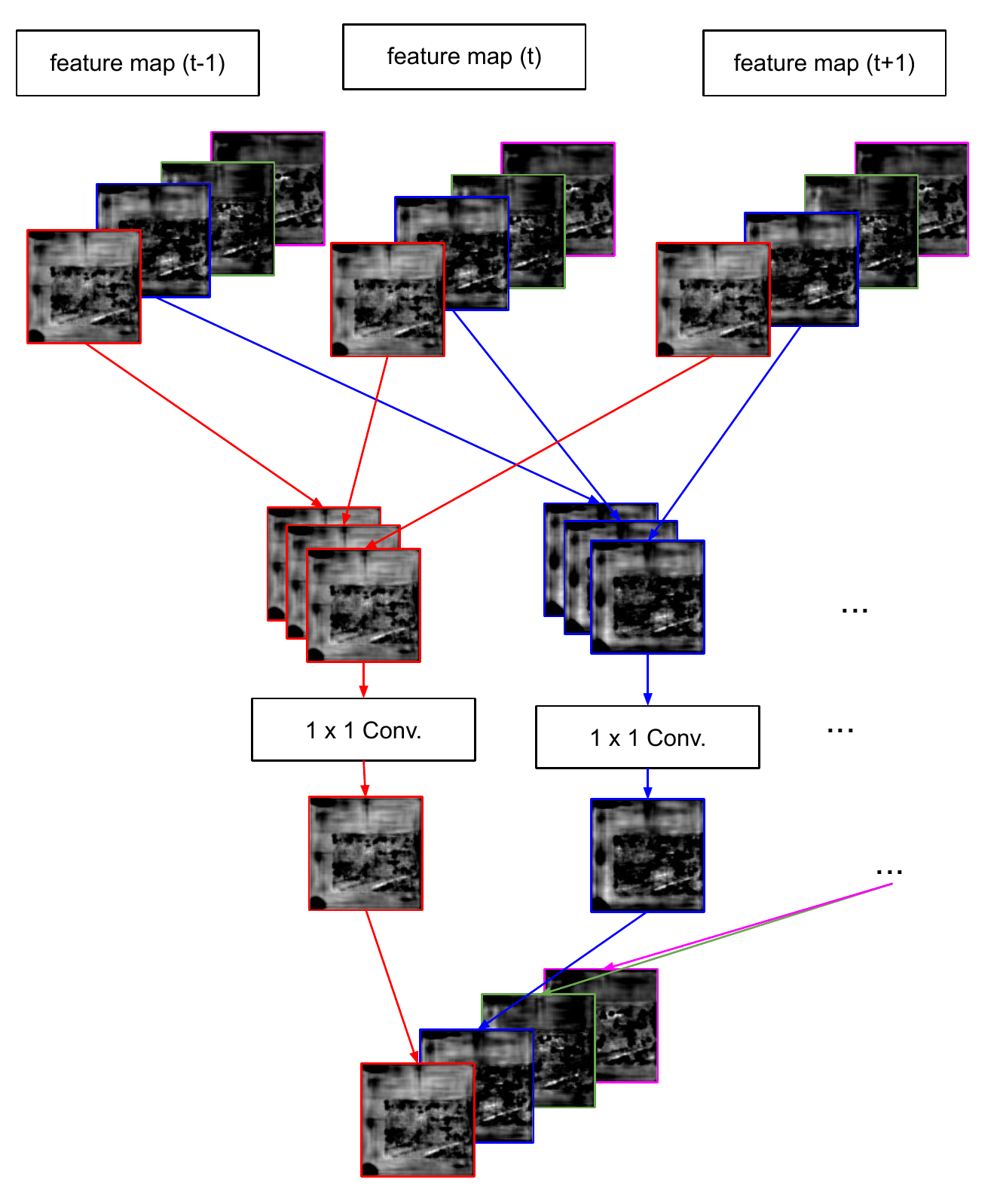}
\vspace{-1em}
\end{center}
   \caption{The fusion module. Channels are represented by colors. The fusion module is composed of channel grouping, concatenation followed by $1 \times 1$ convolution and a final re-ordering of channels.}
\label{fusion}
\vspace{-1em}
\end{figure}

A small trainable module is implemented to merge feature maps of temporally close frames (see figure~\ref{fusion}). The idea for combining feature maps is the intuitive way a human would approach the task. For instance, when you look at one location for a given channel, you might want to average responses over the feature maps from the different frames, or look for the maximum response and only keep this one (like in a max pooling). That would mean the feature would come from the frame where it is best seen. Taking the average responses or the maximum response for merging the feature maps depends on the situation. Therefore, we let the neural network learn the merging operation by itself, using $1 \times 1$ convolutions over the channels.

$1 \times 1$ convolutions are often used to adapt the dimension of feature maps. In GoogLeNet, $1 \times 1$ convolutions are used to reduce the dimension of tensors, allowing the network to be deeper and remaining efficient. Recently, the network BorderDet~\citep{qiu2020borderdet} uses $1 \times 1$ convolutions to learn to produce border sensitive feature maps. In contrast to these previous works, we use $1 \times 1$ convolutions to combine features from several frames. 

The fusion module receives $2n + 1$ feature maps, each of dimension $w * h * c$, as input (see figure~\ref{fusion}). Its output is one merged feature map of dimension $w * h * c$. The feature maps taken as input come from the same backbone networks that share parameters. For $2n + 1$ feature maps of dimension $w * h * c$, we slice every $c$ channels and concatenate them. The result is $c$ tensors of shape $w * h * (2n + 1)$. A two dimensional convolution is performed on these tensors with a kernel of shape $1 \times 1$ ($1 * 1 * (2n + 1)$) and an output depth of 1, resulting in $c$ tensors of shape $w * h$. We finally concatenate the resulting tensors channel-wise to obtain the $w * h * c$ feature map that is our output. This module thus learns the optimal operation to combine feature maps for the domain on which it is fine tuned.

\subsection{SpotNet improvement}
In order to further push the detection accuracy, we propose an improvement to the SpotNet~\citep{perreault2020spotnet} attention module. Instead of using three $3 \times 3$ convolutions to produce the saliency map, we designed and trained a small U-Net segmentation network. The U-Net has four levels and thus reduces the spatial resolution by half four times while doubling the channel resolution four times also, before reversing these changes and returning to the original resolution. This allows the network to produce finer saliency maps and improves detection accuracy.

\section{Experiments}
\subsection{Base object detectors}
\label{basedet}
The  proposed architecture is implemented with several base object detector networks. It was first tested using RetinaNet~\citep{lin2018focal} for its speed and accuracy, along with two other state-of-the-art object detection methods, CenterNet (objects as points)~\citep{zhou2019objectsaspoints} and SpotNet~\citep{perreault2020spotnet}. For the RetinaNet base model, the backbone is a VGG-16 network ~\citep{VGG_Simonyan2014}. For the CenterNet and SpotNet base models, the backbone is a stacked hourglass network~\citep{newell2016stacked}.

For RetinaNet, we merged the three outputs that are used to create the feature pyramid network. The network is the same otherwise. For CenterNet and SpotNet, we merged the outputs of the double stacked hourglass network used as backbone for these detectors. The merged feature map is used to create the center keypoint heatmaps, but the other heads use the target frame feature map. Experiments showed that this worked better than if all the heads used the merged feature maps, maybe due to the fact that the center heatmaps use general spatial features more, and the other heads might be more specialized in some specific semantic features in the feature map that do not answer well to merging between frames. The fusion process is done with our custom fusion module as described above, that enhances the target feature map by merging. Illustrations of the three evaluated models can be found at \protect\url{https://github.com/hu64/FFAVOD}.

\subsection{Datasets}
Since our method relies on temporally close frames, it must be assessed on video datasets. The chosen evaluation domain is traffic surveillance, since it is of great interest to us and there are many applications for video object detection. Two video datasets in the traffic surveillance domain were selected: UA-DETRAC~\citep{Wen2015Tracking} (recorded with a fixed camera) and the Unmanned Aerial Vehicle Benchmark (UAVDT)~\citep{du2018unmanned} (recorded with a mobile camera). The versatility of our architecture is demonstrated by using videos from both fixed and mobile cameras. These two datasets are largely different, UA-DETRAC contains sequences taken by cameras fixed above highways and intersections, with medium sized objects. UAVDT, on the other hand, contain sequences taken by drones that hover over roads at different altitudes, but generally with much higher viewpoints than UA-DETRAC. Therefore the objects are significantly smaller and denser in that dataset, and also the background is changing across the sequence, making it more challenging to merge feature maps. 

\subsection{Implementations Details}
The neural networks are implemented in Keras~\citep{chollet2015keras} using the TensorFlow~\citep{tensorflow2015-whitepaper} backend for our RetinaNet base detector. Our CenterNet and SpotNet base detectors are implemented using Pytorch~\citep{paszke2017automatic_pytorch}. 

The same training protocol was used for the all the base detectors to demonstrate the contribution of our approach. The training process is done in two steps. First, the base detector is fine-tuned on each dataset starting from pre-trained weights on MS COCO~\citep{lin2014microsoft}. Second, the shallower layers of the backbone are frozen and the fusion module as well as the network heads are trained. The reason for freezing the shallower layers is that the network seems to have difficulty training the fusion module while also training every other layers. Doing it in two steps seems to facilitate the learning.

A VGG-16 backbone with a feature pyramid of five levels is used for RetinaNet. RetinaNet takes the outputs of the last three blocks of VGG-16 to build the five-level feature pyramid, three levels of the same dimension of the three VGG-16 blocks, and two smaller. The fusion module is inserted between the VGG-16 and the feature pyramid. As a result, our fusion module is duplicated three times in the network. 

A double stacked hourglass network is used as the backbone network for CenterNet and SpotNet. The fusion module is inserted after the end of the second hourglass. During training, the five frames are therefore passed through the same double stacked hourglass (all the parameters are shared), then passed through the fusion module, and the network continues as usual after that.

To determine the number of frames $n$ used by our model, an ablation study is performed and the results are shown in figure~\ref{ntomap}. We end up using $n=2$, i.e.\ a window of five frames in total, which showed the best performance. 

\subsection{Performance Evaluation}
For the evaluation process, the test set is predetermined on each dataset. The training data is split into training and validation. The split is done by video sequence and not by frame to prevent overfitting on the validation data. The same split of three sets is employed for all of our experiments. We trained by monitoring the validation loss every epoch and select the best model according by the validation loss. Results were then computed on the test set.

The results are evaluated following the dataset protocols, using the Mean Average Precision (mAP). The mAP is the mean of the average precisions for every class. The average precision is the average precision under different recall values, which can also be described as the area under the precision-recall curve. 

\section{Results and Discussion}

\begin{table*}[h!]
\footnotesize
\setlength\tabcolsep{3pt} 
\def\arraystretch{1.5}
\centering
\caption{mAP of FFAVOD applied to base detectors on the UA-DETRAC test set compared their respective base detectors, as well as classic state-of-the-art detectors. FFAVOD uses $n=2$. Results for FFAVOD and their base detectors are generated using the official toolkit from the UA-DETRAC website. \textbf{Boldface} indicates the best result overall, \underline{Underline} indicates the best result within a section, while \textit{Italic} indicates the baseline and *indicates the use of multiple frames.}
\label{resultsuadetrac}
\begin{tabular}{c|c|c|c|c|c|c|c|c|c}
 & Detector & Overall & Easy & Medium & Hard & Cloudy & Night & Rainy & Sunny \\
\hline
\hline
\hline
\multirow{3}{*}{SpotNet} & *FFAVOD-SpotNet with U-Net & \textbf{88.10\%} & \textbf{97.82\%} & \textbf{92.84\%} & \textbf{79.14\%} & \textbf{91.25\%} & \textbf{89.55\%} & \textbf{82.85\%} & 91.72\% \\
\cline{2-10}
& \textit{SpotNet}~\citep{perreault2020spotnet} with U-Net & 87.76\% & 97.78\% & 92.57\% & 78.59\% & 90.88\% & 89.28\% & 82.47\% & \textbf{91.83\%} \\
\cline{2-10}
& SpotNet~\citep{perreault2020spotnet}& 86.80\% & 97.58\% & 92.57\% & 76.58\% & 89.38\% & 89.53\% & 80.93\% & 91.42\% \\
\hline
\hline 
\multirow{2}{*}{CenterNet} & *FFAVOD-CenterNet & \underline{86.85\%} & \underline{97.47\%} & \underline{92.58\%} & \underline{76.51\%} & \underline{89.76\%} & \underline{89.52\%} & \underline{80.80\%} & \underline{90.91\%} \\
\cline{2-10}
& \textit{CenterNet}\citep{duan2019centernet} & 83.48\% & 96.50\% & 90.15\% & 71.46\% & 85.01\% & 88.82\% & 77.78\% & 88.73\% \\
\hline
\hline 
\multirow{2}{*}{RetinaNet} & *FFAVOD-RetinaNet & \underline{70.57\%} & \underline{87.50\%} & \underline{75.53\%} & \underline{58.04\%} & \underline{80.69\%} & \underline{69.56\%}  & \underline{56.15\%} & \underline{83.60\%} \\
\cline{2-10}
& \textit{RetinaNet}~\citep{lin2018focal} & 69.14\% & 86.82\% & 73.70\% & 56.74\% & 79.88\% & 66.57\% & 55.21\% & 82.09\% \\
\hline
\hline
\hline
\multirow{7}{*}{SOTA methods (multiple frames)} & *Joint~\citep{munjal2020joint} & \underline{83.80\%} & - & - & - & - & - & - & -\\
\cline{2-10}
& *Illuminating~\citep{wang2020illuminating} & 80.76\% & \underline{94.56\%} & \underline{85.90\%} & \underline{69.72\%} & \underline{87.19\%} & \underline{80.68\%} & \underline{71.06\%} & \underline{89.74\%}\\
\cline{2-10}
& *MMA~\citep{hu2019mma} & 74.88\% & - & - & - & - & - & - & -\\
\cline{2-10}
& *Global~\citep{lin2021global} & 74.04\% & 91.57\% & 81.45\% & 59.43\% & - & 78.50\% & 65.38\% &  83.53\% \\
\cline{2-10}
& *Perceiving Motion~\citep{liu2019perceiving} & 69.10\% & 90.49\% & 75.21\% & 53.53\% & 83.66\% & 73.97\% & 56.11\% & 72.15\%\\
\cline{2-10}
& *RN-D~\citep{perreault2019road} & 54.69\% &	80.98\% &	59.13\% &	39.23\% &	59.88\% &	54.62\% &	41.11\% &	77.53\% \\
\cline{2-10}
& *3D-DETnet~\citep{3D_detnet_li20183d} & 53.30\% &	66.66\% &	59.26\% &	43.22\% &	63.30\% &	52.90\% &	44.27\% &	71.26\% \\
\hline
\hline
\multirow{7}{*}{SOTA methods (single frame)}& FG-BR\_Net~\citep{fu2019foreground} & \underline{79.96\%} & \underline{93.49\%} & \underline{83.60\%} & \underline{70.78\%} & \underline{87.36\%} & \underline{78.42\%} & \underline{70.50\%} & \underline{89.8\%}\\
\cline{2-10}
& HAT~\citep{wu2019hierarchical} & 78.64\% & 93.44\% & 83.09\% & 68.04\% & 86.27\% & 78.00\% & 67.97\% & 88.78\% \\
\cline{2-10}
& GP-FRCNNm~\citep{amin2017geometric} & 77.96\% & 92.74\% & 82.39\% & 67.22\% & 83.23\% & 77.75\% & 70.17\% & 86.56\% \\
\cline{2-10}
& R-FCN~\citep{RFCN_NIPS2016_6465} & 69.87\% &	93.32\% &	75.67\% &	54.31\% &	74.38\% &	75.09\% &	56.21\% &	84.08\% \\
\cline{2-10}
& EB~\citep{EB_wang2017evolving} & 67.96\%	& 89.65\% &	73.12\% & 53.64\% & 72.42\% & 73.93\% & 53.40\% & 83.73\% \\
\cline{2-10}
& Faster R-CNN~\citep{ren2015faster} & 58.45\% &	82.75\% &	63.05\% &	44.25\% &	66.29\% &	69.85\% &	45.16\% &	62.34\% \\
\cline{2-10}
& YOLOv2~\citep{Redmon_2017_CVPR_YOLO2} & 57.72\% &	83.28\% &	62.25\% &	42.44\% &	57.97\% &	64.53\% &	47.84\% &	69.75\% \\

\end{tabular}
\end{table*}


\begin{table}[h!]
\footnotesize
\setlength\tabcolsep{3pt} 
\def\arraystretch{1.5}
\centering
\caption{mAP of our FFAVOD applied to detectors on the UAVDT test set compared their respective base detectors. FFAVOD uses $n=2$. Results for our FFAVOD and their base detectors are generated using the official Matlab toolkit provided by the authors. The other results are taken from their respective papers. \textbf{Boldface} indicates the best result overall, \underline{Underline} indicates the best result within a section, while \textit{Italic} indicates the baseline and *indicates the use of multiple frames.}
\label{results-UAVDT}
\begin{tabular}{c|c|c}
 & Detector & Overall \\
\hline
\hline
\hline
\multirow{3}{*}{SpotNet} & *FFAVOD-SpotNet with U-Net & \textbf{53.76\%}\\
\cline{2-3}
& \textit{SpotNet}~\citep{perreault2020spotnet} with U-Net& 53.38\%\\
\cline{2-3}
& \textit{SpotNet}~\citep{perreault2020spotnet}& 52.80\%\\
\hline
\hline
\multirow{2}{*}{CenterNet} & *FFAVOD-CenterNet & \underline{52.07\%}\\
\cline{2-3}
& \textit{CenterNet}~\citep{zhou2019objectsaspoints}& 51.18\%\\
\hline
\hline
\multirow{2}{*}{RetinaNet} & *FFAVOD-RetinaNet & \underline{39.43\%}\\
\cline{2-3}
& \textit{RetinaNet}\citep{lin2018focal}& 38.26\%\\
\hline
\hline
\hline
\multirow{5}{*}{SOTA methods} & LRF-NET~\citep{wang2019learning} & \underline{37.81\%}\\
\cline{2-3}
& R-FCN~\citep{RFCN_NIPS2016_6465} &34.35\%\\
\cline{2-3}
& SSD~\citep{liu2016ssd} & 33.62\%\\
\cline{2-3}
& Faster-RCNN~\citep{ren2015faster} & 22.32\%\\
\cline{2-3}
& RON~\citep{kong2017ron} & 21.59\%\\
\end{tabular}
\end{table}

The results on the UA-DETRAC dataset are reported in table~\ref{resultsuadetrac}. The proposed FFAVOD applied to the three base detectors consistently outperforms them. The FFAVOD applied on SpotNet achieves the state-of-the-art (SOTA) result on this dataset. Our approach improves the performance of the base detectors particularly well on harder categories like ``hard'' and ``rainy'', hinting that indeed our FFAVOD allows the network to overcome challenges present on the harder examples, but has less impact on the easier ones.
There is a significant improvement of the detection results when using our FFAVOD with the RetinaNet detector as well as on the CenterNet detector. It is also true for SpotNet, but the improvement is smaller. This shows that our proposed feature fusion architecture can capitalize on multiple frames to improve object detection, and that our approach is applicable to several networks. In general, the performance is never reduced when using FFAVOD, except in the ``Sunny'' case for SpotNet where the mAP is virtually identical, due to the examples being easy. Even though they achieve good results, no other detector using multiple frames can outperform the results that we obtained with modern detectors (FFAVOD-CenterNet and FFAVOD-SpotNet). This shows that our approach capitalizes better on the multiple frames since it can be integrated with SOTA single frame detectors to improve them. Also, our SpotNet with U-Net outperforms the original SpotNet by around 0.5 to 1\% in both datasets, showing the significance of using a better attention module.  

The results on the UAVDT dataset are reported in table~\ref{results-UAVDT}. Our FFAVOD detectors consistently outperform their respective base detectors. On the SpotNet architecture, the FFAVOD achieves also the state-of-the-art result on this dataset.

The improvement is smaller on SpotNet. Our hypothesis for why that is is that the attention module of SpotNet might help detect many of the same objects (small or occluded) that our fusion module does. Nevertheless, the improvement is still significant. On UA-DETRAC the improvement is almost zero for the ``Easy'' category, but close 1\% for the ``Hard'' category, which is very consistent with the idea that the fusion module helps overcome challenges on hard examples. On UAV, the improvement is consistent with UA-DETRAC. It is also important to note that the better we perform on a dataset, the harder it becomes to improve our results.

\subsection{Ablation Study}

An ablation study was designed to evaluate the contribution of the different parts of the proposed FFAVOD. The two contributions were isolated and analyzed. On the UA-DETRAC dataset using SpotNet, the fusion module was removed and replaced with a simple concatenation followed by a convolution. Instead of channel by channel grouping and convolution then reordering, we simply concatenate the feature maps and reduce the dimension with a convolution. Replacing the fusion module with concatenation decreased the performance by a large margin as shown in table~\ref{results-ablation} (Concatenation). We believe that combining feature maps in this fashion is noisy and the model might require more parameters in order to learn how to combine them. The fusion module on the other hand is a much more direct and efficient way of combining feature maps. As baseline fusion methods, we also tried using a mean, maximum and a median operation, and we show that the results are worse than the single frame method. Additionally, to justify the use of past and future frames, we did an experiment where we retrain our model using $2n$ past frames with the target frame for $n = 1$ and $n = 2$. This decreased performance dramatically, as the fusion module has a hard time focusing on the target frame and the positions of the objects get too different (features are not aligned). Indeed, the distribution of the spatial information is no longer centered around the target frame, and this causes misalignment of the bounding boxes.  

\begin{figure}[t]
\begin{center}
\includegraphics[width=0.8\linewidth]{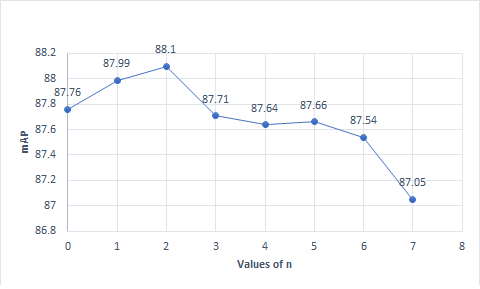}
\vspace{-1em}
\end{center}
   \caption{mAP reported on the UA-DETRAC test set for different values of $n$ of FFAVOD}
\label{ntomap}
\vspace{-1.5em}
\end{figure}

In order to select the number of frames, FFAVOD was applied by varying the number of frames used by the network (see figure~\ref{ntomap}). Each model using a particular $n$ is re-trained by freezing the base network weights and training only the fusion module weights. We show that we obtain the highest result by using 5 frames, and thus that is what we used to produce our main results.  

\begin{table}[h!]
\footnotesize
\setlength\tabcolsep{3pt} 
\def\arraystretch{1.5}
\centering
\caption{Different fusion strategies to conduct an ablation study. Results are generated using the official Matlab toolbox provided by the authors. \textbf{Boldface} indicates the best result overall}
\label{results-ablation}
\begin{tabular}{c|c|c}
$n$ & Fusion Method & mAP \\
\hline
\hline
2 & Learned (ours, as proposed) & \textbf{88.10\%} \\
\hline
0 & None (baseline) & 87.76\% \\
\hline
2 & Max & 87.09\% \\
\hline
2 & Mean & 87.09\% \\
\hline
2 & Median & 87.08\%\\
\hline
2 & Concatenation & 83.44\%\\
\hline
2 & Learned (ours) (past frames only) & 75.20\%\\
\hline
1 & Learned (ours) (past frames only) & 76.02\%\\
\end{tabular}
\end{table}
\subsection{Limitations}
One obvious limitation of our architecture is the fact that it needs video sequences with temporally close frames in order to work. This limitation is mitigated by the fact that a lot of applications for object detection rely on such data. Another limitation is the memory usage increase in inference, where FFAVOD has to keep feature maps stored in memory for several iterations before releasing them when they are no longer in the target frame window. This makes our architecture not ideal for embedded applications. Finally, without some sort of tracking, the temporal scope is also somewhat limited, making bigger improvements in accuracy difficult.  

\section{Conclusion}
We introduce FFAVOD, a new method for video object detection which can be applied and used with most standard object detectors. Using the proposed approach, we trained and evaluated our fusion module on two datasets from the traffic surveillance domain. We demonstrate with three different base detectors that performance can be significantly increased by helping solve challenges in the harder examples of the datasets. Additionally, we propose an improvement to the SpotNet attention module that increases the detection accuracy. Several ideas may be tried to further increase performance, for example by adding temporal coherence in the form of re-identification of features across frames, or by integrating tracking in parallel to detection.

\section{Acknowledgment}
\vspace{-0.2em}
We acknowledge the support of the Natural Sciences and Engineering  Research Council of Canada (NSERC), [RDCPJ 508883 - 17], and the support of Genetec.

\bibliographystyle{model2-names}
\bibliography{bib}

\end{document}